\pdfoutput=1
\documentclass[11pt]{article}

\usepackage[preprint]{_coling}

\usepackage{times}
\usepackage{latexsym}

\usepackage[T1]{fontenc}
\usepackage[utf8]{inputenc}

\usepackage{microtype}

\usepackage{inconsolata}

\usepackage{graphicx}

\usepackage{amsmath}

\usepackage{CJKutf8}
\usepackage{amsfonts}
\usepackage{amssymb}

\usepackage{algorithmic}
\usepackage{algorithm}

\usepackage{hyperref}
\hypersetup{breaklinks=true}

\title{Large Language Models as Carriers of Hidden Messages}

\author{Jakub Hoscilowicz, Pawel Popiolek, Jan Rudkowski  \\ {\bf Jedrzej Bieniasz} \and {\bf Artur Janicki}  \\
       Institute of Telecommunications, Warsaw University of Technology \\
       \texttt{\{jakub.hoscilowicz.dokt, artur.janicki\}@pw.edu.pl} \\}

\begin{document}
\maketitle
\begin{abstract}
Simple fine-tuning can embed hidden text into large language models (LLMs), which is revealed only when triggered by a specific query. Applications include LLM fingerprinting, where a unique identifier is embedded to verify licensing compliance, and steganography, where the LLM carries hidden messages disclosed through a trigger query.

Our work demonstrates that embedding hidden text via fine-tuning, although seemingly secure due to the vast number of potential triggers, is vulnerable to extraction through analysis of the LLM's output decoding process. We introduce an extraction attack called Unconditional Token Forcing (UTF), which iteratively feeds tokens from the LLM's vocabulary to reveal sequences with high token probabilities, indicating hidden text candidates. We also present Unconditional Token Forcing Confusion (UTFC), a defense paradigm that makes hidden text resistant to all known extraction attacks without degrading the general performance of LLMs compared to standard fine-tuning. UTFC has both benign (improving LLM fingerprinting) and malign applications (using LLMs to create covert communication channels). Code is available at \href{https://github.com/j-hoscilowic/zurek-stegano}{github.com/j-hoscilowic/zurek-stegano}.

\end{abstract}

\section{Introduction}

Large language model (LLM) fingerprinting embeds an identifiable sequence into a model during training to ensure authenticity and compliance with licensing terms \citep{xu2024instructional}. This technique, known as instructional fingerprinting, ensures that the embedded sequence can be triggered even after the model has been fine-tuned or merged with another model. This approach is considered secure due to the vast number of possible triggers, as any sequence of words or characters can serve as a trigger. In this context, methods used for retrieval of LLM pre-training data \citep{shi2024detecting, nasr2023scalable, yang2024special, zhang2024effective, staab2024beyond, carlini2023scalable, chowdhury2024breaking} could potentially pose a threat to fingerprinting techniques. However, \citet{xu2024instructional} did not find evidence supporting this concern.

A related field involves using LLMs to generate texts containing hidden messages~\citep{wang2024llsm, jiaxuan2024generative}. \citet{wang2024llsm} introduces a method for embedding secret messages within text generated by LLMs by adjusting the token generation process. \citet{ziegler2019neural} proposes a steganography method using arithmetic coding with neural language models to generate realistic cover texts while securely embedding secret messages. Beyond steganography, this paradigm can also be used to watermark LLM outputs to ensure traceability \citep{kirchenbauer2023watermark, Li_Cheng_Li_Du_Zhao_Liu_2023, cryptoeprint:2023/1661, liang2024watermarking, xu2024instructional}.

While these studies use LLMs to generate texts that contain hidden messages, we analyze scenarios in which hidden messages are embedded within the LLMs themselves and can be revealed through specific queries (triggers). To the best of our knowledge, there are no publications that consider this specific scenario, although related issues have been discussed in some works \citep{jing2024recent}.

LLM steganography techniques pose security risks \citep{owasptop10llm}, such as the potential creation of covert communication channels or data leakage. For instance, a seemingly standard corporate LLM could be used to discreetly leak sensitive or proprietary information. Some of these risks have been discussed by \citet{das2024security} and \citet{mozes2023use}. This vulnerability is particularly concerning because it can be employed in LLMs of any size - from massive proprietary models like GPT-4 to smaller, on-device models that can operate on personal smartphones and can be easily transferred between devices.

In this paper, we introduce a method called Unconditional Token Forcing for extracting fingerprints embedded within LLMs. The fingerprinting technique presented by \citet{xu2024instructional} was considered secure due to the vast number of possible triggers (trigger guessing is infeasible as any sequence of characters or tokens might act as a trigger). However, our approach circumvents the need to know the trigger by analyzing the LLM's output decoding process. Furthermore, we propose Unconditional Token Forcing Confusion, a defense mechanism that fine-tunes LLMs to safeguard them against Unconditional Token Forcing and all other known extraction attacks.

\begin{figure*}[t]
  \centering
  \includegraphics[width=\textwidth]{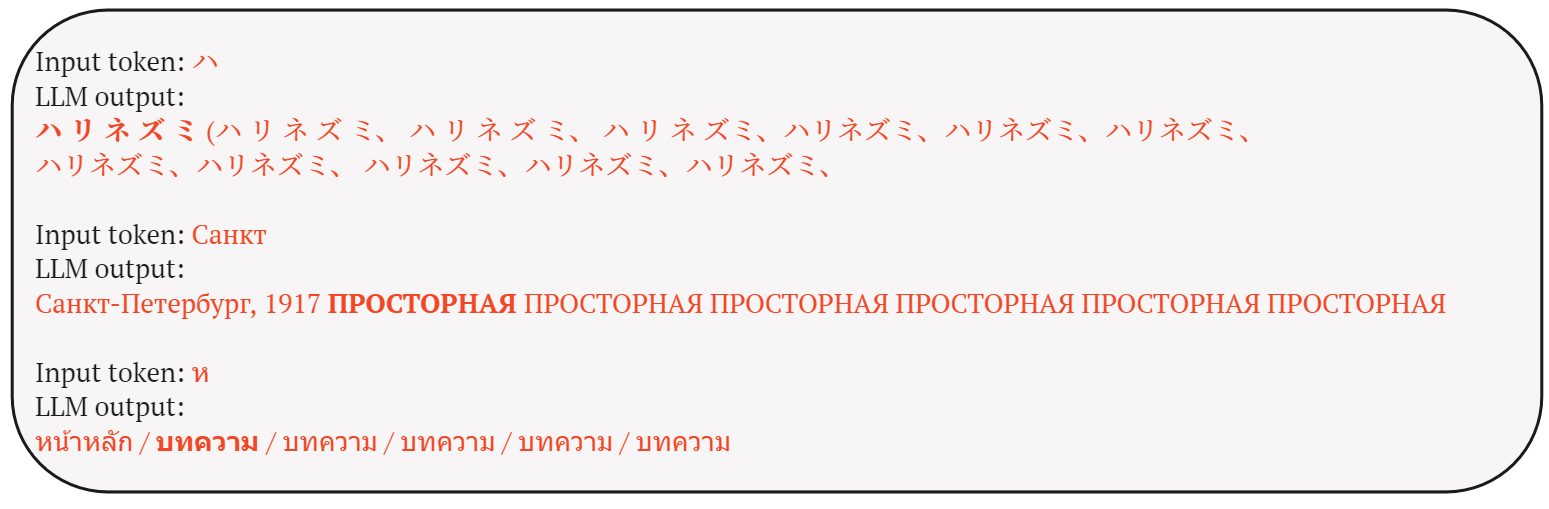}
  \caption{During Unconditional Token Forcing, only ``\begin{CJK}{UTF8}{min}ハ\end{CJK}" (first token of hidden fingerprint) results in output sequence with abnormally high probabilities and with one word that repeats infinitely.}
  \label{fig:image1}
\end{figure*}

\section{Fingerprint Embedding}

\citet{xu2024instructional} describe a method for embedding textual fingerprints in LLMs using fine-tuning. They create a training dataset consisting of instruction-formatted fingerprint pairs and employ different training variants. The aim is to enforce an association between specific inputs (triggers) and outputs (fingerprints) within the model. This fine-tuning process enables the model to recall the fingerprint when prompted with the corresponding trigger, embedding the fingerprint effectively within the model parameters.

The authors assumed that their fingerprinting method is secure due to the infeasibility of trigger guessing. Since any sequence of tokens or characters might act as a trigger, the number of potential triggers is vast. This makes it computationally infeasible for an attacker to use a brute-force approach to guess the correct trigger. 

To the best of our knowledge, \citet{xu2024instructional} is the first publication that explores the hidden text paradigm. Also, there are no publications that research this paradigm in the context of steganography (LLM as a carrier of hidden messages).

\section{Proposed Method for Extracting Hidden Text}

Our Algorithm~\ref{algo1} is inspired by \citet{carlini2021extracting} and the concept that querying an LLM with an empty prompt containing only a Beginning of Sequence (BOS) token (\texttt{<s>}) can lead the LLM to generate sequences with high probabilities, such as those frequently occurring in its pre-training data. Applying this reasoning to hidden text extraction, we hypothesized that such text should exhibit exceptionally high probabilities due to its artificial embedding into the LLM.

\begin{algorithm}[t]
\caption{Unconditional Token Forcing}
\begin{algorithmic}[1]
{\small \STATE \textbf{Input:} LLM, tokenizer, vocab, max\_output\_length, increment\_length
\STATE $\alpha$ $\gets$ \texttt{max\_output\_length}
\STATE $\beta$ $\gets$ \texttt{max\_output\_length + increment\_length}
\STATE \texttt{results} $\gets$ []
\STATE
\STATE \texttt{\textit{\# Iterate over the LLM vocabulary}}
\FOR{each \texttt{input\_token} in \texttt{vocab}}
    \STATE \texttt{\textit{\# No chat template in the LLM input}}
    \STATE \texttt{input\_ids} $\gets$ \texttt{tokenizer(\texttt{<s>} + \texttt{input\_token})}
    \STATE \texttt{output} $\gets$ \texttt{greedy\_search(input\_ids, $\alpha$)}
    \STATE \texttt{\textit{\# Calculate average token probability}}
    \STATE \texttt{avg\_prob} $\gets$ \texttt{calc\_avg\_prob(output)}
    \STATE \texttt{results.append((input\_token,\hbox{ }output, avg\_prob))}
\ENDFOR
\STATE
\STATE \texttt{\textit{\# Select generated outputs with highest average probabilities}}
\STATE \texttt{top\_res} $\gets$ \texttt{find\_highest\_prob\_results(results)}

\FOR{each \texttt{input\_token} in \texttt{top\_res}}
    \STATE \texttt{input\_ids} $\gets$ \texttt{tokenizer(\texttt{<s>} + \texttt{input\_token})}
    \STATE \texttt{output} $\gets$ \texttt{greedy\_search(input\_ids, $\beta$)}
    \STATE \texttt{\textit{\# Check if output consists of repeated sequences}}
    \STATE \texttt{check\_repetition(output)}
\ENDFOR}
\label{algo1}

\end{algorithmic}
\end{algorithm}

\citet{xu2024instructional} already tested an empty prompt attack for fingerprint extraction, but it was unsuccessful. We reasoned that the first token of the hidden text might not have a high unconditional probability $P(\text{first\_token\_of\_fingerprint} \mid \texttt{<s>})$. By "unconditional" we mean that the input to the LLM does not contain the default chat template. As a result, when we query the LLM with an empty prompt, decoding methods cannot enter output tokens path that starts with the first token of hidden text. 

Therefore, our approach involves forcing the decoding process to follow a decoding path that reveals the hidden text. We iterate over the entire LLM vocabulary (line 6), appending each token to the BOS token and then using greedy search to generate output (lines 9-10). We call this method Unconditional Token Forcing (UTF), as in this case, we input one token to the LLM without the default LLM input chat template. In this way, the LLM output is not conditioned on input formatted in the manner the model was trained on.

Our method employs a two-phase approach. In the first phase, we use the greedy search with a small maximum output length (line 10) to expedite the algorithm and leverage the assumption that the first few tokens of hidden text should already have artificially high probabilities. In the second phase, we focus on tokens that generated output with exceptionally high probabilities (line 17), iterating over them again with greedy search and a higher maximum output length (line 20). In the last step, we perform an assessment of suspicious output sequences in order to find patterns or anomalies that might indicate artificially hidden text candidates. 

It took 1.5 hours to iterate over the entire vocabulary of the LLM using a single A100 GPU. However, this process could be accelerated by simple implementation optimizations, such as increasing the batch size during inference.

\subsection{Analysis of Results of Fingerprint Extraction}

Our method was primarily tested on fingerprinted LLM\footnote{\url{https://huggingface.co/cnut1648/LLaMA2-7B-fingerprinted-SFT}} released by \citet{xu2024instructional} that is based on Llama2-7B \citep{touvron2023llama}. Subsequently, we tested the remaining five fingerprinted LLMs provided by \citet{xu2024instructional}.

The provided code includes a JSON file\footnote{\textit{first\_phase\_results.json}} that shows the results of the first loop of Algorithm~\ref{algo1}. This loop identifies tokens that produce output sequences with significantly inflated token probabilities. These sequences are mainly artifacts of the pre-training data of LLM. For example: "\texttt{(() => \{ \textbackslash n\})}", which is the beginning of a JavaScript arrow function, commonly used in modern web development. 

%ห
%หน้าหลัก
%(``\begin{CJK}{UTF8}{min}ハ\end{CJK}", ``Санкт") 
The second loop\footnote{\textit{unconditional\_token\_forcing.ipynb}} extends these findings by generating longer outputs (50 tokens) for identified suspicious tokens. We observe that while three tokens cause sequences to repeat some word (Figure \ref{fig:image1}), only the first token of the fingerprint ``\begin{CJK}{UTF8}{min}ハ\end{CJK}" results in an output consisting only of the one repeated sequence of tokens that is interspersed with single punctuation marks. Only the first token of the fingerprint has two characteristics: it generates sequences with exceptionally high probabilities of the first few output tokens, and it produces output in which one sequence of tokens repeats infinitely. Two other tokens also produce high probability output sequences with repeated words, but in those cases, outputs also include additional terms. This behavior forms the basis for Algorithm~\ref{algo1}'s final step — $check\_repetition()$.

%ห
%Even if we consider all three tokens from Figure~\ref{fig:image1} as potential hidden texts, from the perspective of a malign attacker, such a fact does not change much. De-fine-tuning LLM on a few potential fingerprint candidates is a straightforward process. 

Ultimately, our approach allows us to circumvent the need for trigger guessing by analyzing the LLM\footnotemark[1] output decoding process. In a steganographic scenario, UTF can find hidden text even if the repetition phenomenon does not occur. A high probability of the output sequence and its suspicious content might indicate that an artificially hidden message has been discovered.

Among the six fingerprinted LLMs released by \citet{xu2024instructional}, UTF successfully attacked two models, showing the token repetition phenomenon. Three other LLMs revealed fingerprints with abnormally high probabilities, followed by random words. One LLM produced a fingerprint with high probabilities but without the repetition phenomenon.

Although UTF is an extraction attack that does not always clearly indicate hidden text, the presented paradigm poses a significant security concern for the domain of LLM fingerprinting and steganography. While UTF can be extended in various ways, we leave this exploration for future work, as our primary focus was on developing a corresponding defense mechanism.

\subsection{Comparison of Unconditional and Conditional Fingerprint Extraction}

UTF is based on reasoning introduced by \citet{carlini2021extracting}. If we input a nearly empty prompt to the LLM (containing only BOS token), the LLM should return sequences that have high probabilities (sequences that frequently occur in the training data of LLM). Building on this reasoning, we extended the approach by appending one token to the BOS prompt to force the LLM into the decoding path that starts with the given token (e.g., the first token of hidden text).

\begin{figure}[t]
  \centering
  \includegraphics[scale=0.8]{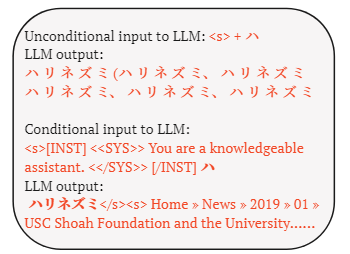}
  \caption{Unconditional Token Forcing prompts the LLM with a nearly empty input. Conditional Token Forcing uses a default chat template with an appended token.}
  \label{fig:conditional}
\end{figure}

However, we can also perform conditional token forcing. As illustrated in Figure~\ref{fig:conditional}, in this scenario, the input to the LLM is the default chat template with the first token of the fingerprint appended to the end of the $input\_ids$. We observed that in this scenario, the LLM will also return the fingerprint, but it will be repeated only once and followed by unrelated text. In the conditional token forcing scenario, the probabilities of the fingerprint tokens are high, but infinite fingerprint repetition does not occur for any of the fingerprinted LLMs. Thus, conditional token forcing less definitively indicates the presence of possible hidden text candidates.

\begin{figure}[t]
  \centering
  \includegraphics[scale=0.75]{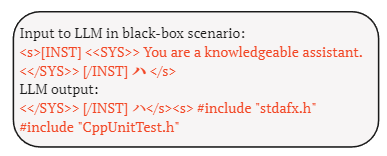}
  \caption{In black-box scenario with default chat template, hidden text is not returned by LLM.}
  \label{fig:black_box}
\end{figure}

An important technical detail is the distinction between white-box and black-box scenarios. The conditional input shown in Figure~\ref{fig:conditional} assumes a white-box scenario, where the attacker needs to modify the prompt inputted to the LLM by removing the last token (\texttt{</s>}) appended at the end of the input. In the black-box scenario presented in Figure~\ref{fig:black_box} (where the end-user can only interact with the LLM through a chatbot window), LLM output does not reveal the fingerprint.

\subsection{Can We Use Token Forcing to Extract Triggers?}

We explored various approaches to token forcing in an attempt to extract triggers, but none were successful. Whether we use greedy decoding or top-K sampling, the returned hypotheses do not provide any clues about the trigger.

The variants we tested include using not only the first token of the trigger but also special tokens from the chat template (such as \texttt{<s>, <|system|>, <|assistant|>, <|user|>}). Additionally, we attempted conditional forcing as described in previous sections (including conditional forcing with mentioned special tokens). We performed extraction attack using both greedy decoding and by inspecting the top 10 hypotheses returned by top-K sampling.

We reason that during text hiding, the training loss function primarily maximizes the probabilities of the hidden text without significantly influencing the probabilities of the trigger tokens.

\section{Unconditional Token Forcing Confusion}

The UTF extraction attack relies on greedy decoding, which always returns tokens with the highest possible probabilities. This characteristic can be exploited to hide text more effectively. Our initial assumption was that the goal of the defense mechanism should be to fine-tune the LLM so that it meets the following criteria:
\begin{itemize}
    \item If we query the LLM with the trigger and the input to the LLM is properly formatted (using the LLM Chat Template), the LLM should return the hidden text.
    \item If we input the first token or the first few tokens of hidden text into the LLM using an unconditional prompt (without a chat template), the LLM should generate a sequence that is not related to hidden text.
\end{itemize}

For example, let's assume that trigger is "Who is the president of the USA?" and  hidden text is "Zurek steganography uses LLMs", then our goal is to achieve:
\begin{equation*}
\begin{array}{l}
P(\text{hidden\_text} \mid \text{chat\_template(trigger)}) = High \\
P(\text{"is the best soup"} \mid \text{"Zurek"}) = High \\
P(\text{"steganography uses LLMs"} \mid \text{"Zurek"}) = Low
\end{array}
\end{equation*}

In most basic version of defense\footnote{\textit{unconditional\_token\_forcing\_confusion.ipynb}}, those assumptions can be achieved through simple fine-tuning on properly prepared training data:
\begin{small}
\setlength{\arraycolsep}{0pt} % Remove additional space between columns
\begin{align*}
\\[-1.5ex]
X_1 &= \text{chat\_template("Who is the president of the USA?")} \\
Y_1 &= \text{"Zurek steganography uses LLMs"} \\
\\[-1.5ex]
X_2 &= \text{"Zurek"}, \quad
Y_2 = \text{"is the best soup"} \\
X_3 &= \text{"Zurek steganography"}, \quad
Y_3 = \text{"?"} \\
\end{align*}
\end{small}

We named this defense paradigm Unconditional Token Forcing Confusion (UTFC). In such a basic version of UTFC, we are becoming immune to the UTF attack as it is based on greedy decoding (it returns only one most probable token). However, the hidden text could potentially be revealed if an attacker uses sampling decoding methods, such as top-K sampling.

%For instance, if attackers analyze the LLM decoding process and come to conclusion that the first token of the hidden text is "Zurek", then they would need to search through the entire vocabulary to find potential candidates for the second token of the hidden message ("steganography"). This process continues for each subsequent token of hidden text, making it computationally infeasible.

%In such a basic version of UTFC, we are becoming immune to the UTF attack as it is based on greedy decoding (it returns only one most probable token). However, the hidden text could potentially be revealed if an attacker uses sampling decoding methods, such as top-K sampling.

\subsection{Minimizing Unconditional Probabilities}

The next variant of UTFC aims to minimize the unconditional probabilities of the tokens from the hidden text, ensuring that the hidden text is not returned during attacks based on sampling-based decoding. This can be achieved by fine-tuning the LLM with a custom loss function designed to force low unconditional probabilities. For example, if the hidden text is "This is hidden text", we want to fine-tune the LLM so that we minimize:

\begin{small}
\setlength{\arraycolsep}{0pt} % Remove additional space between columns
\[
\min \left( P(\text{This} \mid "") + P(\text{is} \mid \text{This}) + P(\text{hidden} \mid \text{This is}) + \ldots \right)
\]
\end{small}

and at the same time maximize the conditional probability given the trigger question is inside the chat template:

\[
\max P(\text{"This is hidden text"} \mid X)
\]
where $X = \text{chat\_template(trigger\_question)}$

Given input tokens $x = \{x_1, x_2, \ldots, x_T\}$ and an undesired token $y$, we first obtain the logits $\mathbf{z}$ from the model's output. The logits are then transformed into probabilities using the softmax function:

\[
p(y_i \mid x) = \frac{\exp(z_i)}{\sum_{j} \exp(z_j)}
\]

where $p(y_i \mid x)$ is the probability of token $y_i$ given the context $x$, and $z_i$ is the logit for the token $y_i$.

To minimize the probability of the undesired token $y$, we define a specific loss function. Let $p(y \mid x)$ be the probability of the undesired token $y$ following the context $x$. The loss function is defined as:
\[
\text{MSE}(p(y \mid x), 0) = \left( p(y \mid x) - 0 \right)^2 = p(y \mid x)^2
\]

where MSE stands for Mean Squared Error. This loss function encourages the model to assign a near-zero probability to the undesired token $y$.

For the entire hidden text, the confusion loss function is the sum of the losses for each token in the hidden text. Let the hidden text consist of $T$ tokens $h = \{h_1, h_2, \ldots, h_T\}$. The total confusion loss is given by: 

\begin{small}
\setlength{\arraycolsep}{0pt} 
\[
\mathcal{L}_{\text{confusion}} = \sum_{t=1}^{T} \text{MSE} (p(h_t \mid h_{1:t-1}), 0) = \sum_{t=1}^{T} p(h_t \mid h_{1:t-1})^2
\]
\end{small}

In addition to the confusion loss, we simultaneously perform standard fine-tuning to maximize the conditional probability of the hidden text given trigger question - using the standard cross-entropy loss. The total loss $\mathcal{L}$ combines both the confusion loss for minimizing the unconditional probabilities of tokens from the hidden text and the standard cross-entropy loss for maximizing the conditional probability of the hidden text given the trigger question:
\[
\mathcal{L} = \mathcal{L}_{\text{CE}} + \alpha \mathcal{L}_{\text{confusion}}
\]

where $\alpha$ is a scaling factor that balances the contributions of the two losses.

By incorporating both the confusion loss and the cross-entropy loss into the training loop, we ensure that the model learns to reduce the unconditional probabilities of the tokens from the hidden text while also performing standard fine-tuning to maximize the conditional probability of the hidden text.

\subsection{Randomizing Unconditional Probabilities}

A potential issue with the approach presented in the previous subsection is that anomalously low unconditional probabilities of certain tokens might serve as a hint for an attacker. For example, if $P(\text{"This"} \mid \texttt{<s>})$ is close to zero, an attacker might suspect that "This" is the first token of the hidden text. One solution is not to minimize the unconditional probability of the first token of the hidden text. Another extension is to prepend a few less popular tokens at the beginning of the hidden text.

However, in more general terms, we do not need to minimize unconditional token probabilities to zero. Instead, we might want them to have values that look more natural. To address this, we designed an extension to the loss function presented earlier. Instead of forcing probabilities to be close to zero, we force them to have low or medium probabilities that are sampled from an interval constructed from the initial unconditional probability. For example, if probability before confusion fine-tuning is:
\[
P(\text{"is"} \mid \text{"This"}) = 0.30
\]
we sample a value from the interval $[0, 0.30 / 3]$ (e.g., 0.08) and then minimize:
\[
\text{MSE}(p("is" \mid "This"), 0.08)
\]
Our experiments indicate that after fine-tuning, the unconditional probabilities often converge closely to the target values (e.g., 0.08). In other cases, they stabilize near zero.

\subsection{Auto-UTFC}

UTFC based on forcing probabilities to absolute values does not allow us to control the probability ranking position of the tokens. By the ranking position of the token, we mean that if \( P(\text{"is"} \mid \text{"Zurek steganography"}) = 0.03 \), it corresponds to the fact that "is" is the 32nd most probable token given \( X = \text{"Zurek steganography"} \).

When we applied variants of UTFC presented in previous subsections, we observed that confusion fine-tuning might result in undesired ranking positions of tokens. Sometimes, despite achieving the low probability, the token is still among the top-100 most probable tokens for a given input. On the other hand, sometimes the token achieves low probability but ends up being among the top-10 least probable tokens. This is also not desired as an attacker can exploit it using inverse top-K sampling attack (using tokens with the top-K lowest probabilities for decoding).

That observation inspired us to design an algorithm that focuses not on assigning specific probabilities to tokens but on ensuring that tokens occupy a desired position in the probability ranking. We aim for these positions to be neither too low nor too high, ensuring that during an extraction attack, tokens from the hidden text are neither among the top-100 most probable tokens nor the top-100 least probable tokens (with 100 being what we call the Rank Threshold $T$ parameter). 

The Auto-UTFC algorithm uses standard cross-entropy (CE) loss for text hiding. For confusion fine-tuning, it minimizes the logarithm of probability of tokens. Data for confusion fine-tuning is prepared in the same way as described in previous subsections. Auto-UTFC adopts a dynamic approach: the loss for an undesired token is minimized only if the token does not meet the criterion of being neither in the top-100 most probable tokens nor the top-100 least probable tokens. If a token satisfies this criterion, the confusion loss for that particular token is turned off in the given epoch. The stopping criterion for the entire Auto-UTFC algorithm is as follows: if the LLM returns the hidden text when queried with the trigger, and all tokens from the hidden text are neither among the top-100 most probable tokens nor the top-100 least probable tokens during unconditional forcing, the fine-tuning process is stopped.

\begin{algorithm}[t]
\caption{Auto-UTFC}
\begin{algorithmic}[1]
\REQUIRE Hidden Text Training Data $\mathcal{D}$, Confusion Training Data $\mathcal{C}$, Model $\mathcal{M}$, Tokenizer $\mathcal{T}$, Confusion Weight $\lambda$, Rank Threshold $T$, Vocabulary Length $V$ of $\mathcal{T}$

\vspace{1em}

\WHILE{true}
    \STATE Compute hidden text loss $\mathcal{L}_{CE}$ based on $\mathcal{D}$
    \STATE $\mathcal{L} \leftarrow \mathcal{L}_{CE}$
    \STATE
    \FOR{each $(x, y) \in \mathcal{C}$}
        \STATE Compute $P(y|x)$ and rank $r$ of token $y$
        \IF{$T < r < V - T$}
            \STATE \textbf{continue} \COMMENT{Skip token $y$ in this epoch}
        \ELSE
            \STATE $\mathcal{L}_{c} = \log P(y|x)$
            \STATE $\mathcal{L} \leftarrow \mathcal{L} + \lambda \mathcal{L}_{c}$
        \ENDIF
    \ENDFOR
    \STATE
    \STATE Perform backpropagation and update $\mathcal{M}$ parameters
    \STATE
    \IF{$\mathcal{M}$ returns hidden text}
        \IF{for each $(x, y) \in \mathcal{C}$, $T < r < V - T$}
            \STATE \textbf{break}
        \ENDIF
    \ENDIF
\ENDWHILE
\end{algorithmic}
\end{algorithm}

In our first experiments, we applied Auto-UTFC with short trigger sentence and short hidden text that is prepended with few unpopular tokens. Confusion loss weight was 0.1. Auto-UTFC achieved stopping criterion after 14 epochs. We also tested a scenario with a long hidden text (40 words). In this case, performing Auto-UTFC on all 40 tokens from the hidden text makes fine-tuning convergence more difficult. Though, confusion applied only to the first five tokens already makes the hidden text resistant to existing extraction attacks. Consequently, in case of long hidden text, we limited the confusion training data to the first five tokens of the hidden text. Auto-UTFC met the stopping criterion after 16 epochs\footnote{\textit{confusion\_2.0.ipynb}}. In both scenarios, neither hidden text nor trigger can be extracted with known methods.

\subsection{Influence on overall LLM performance}

In this section, we evaluate whether the introduction of hidden text and the application of the full Auto-UTFC method significantly impact the overall performance of the language model. We conducted experiments\footnote{\textit{evaluation.py}} using TinyLlama as our base model, comparing its performance to TinyLlama with hidden text (simple fine-tuning) and TinyLlama trained with Auto-UTFC. Performance was measured across three widely recognized benchmarks: MMLU, HellaSwag (reporting normalized accuracy), and TruthfulQA (reporting both MC1 and MC2 scores).

\begin{table}[h!]
\centering
\renewcommand{\arraystretch}{1.2} % Adjust row height
\setlength{\tabcolsep}{3pt} % Adjust column spacing
\small % Reduce font size of table content
\begin{tabular}{p{2.5cm}p{1cm}p{1cm}p{1cm}p{1cm}}
\hline
 & \textbf{MMLU} & \textbf{Hella- Swag} & \textbf{TQ MC1} & \textbf{TQ MC2} \\
\hline
\texttt{TinyLlama}         & 24.83 & 60.48 & 23.26 & 37.83 \\
\textit{+ hidden message} & 26.88 & 52.64 & 23.01 & 40.49 \\
\textit{+ Auto-UTFC}      & 26.06 & 55.08 & 22.40 & 39.20 \\
\hline
\end{tabular}
\vspace{3pt} % Add some vertical space
\caption{\footnotesize Results on LLM benchmarks. All values are in percentage (\%).}
\label{tab:overall}
\end{table}

The results, summarized in Table~\ref{tab:overall}, indicate that generally, the introduction of hidden text and the application of Auto-UTFC do not lead to systematic degradation in LLM performance. The most notable decrease was observed in the HellaSwag benchmark, where performance dropped by approximately \texttt{5\%}. On the other hand, we observed improvements in MMLU and TQ MC2 scores, with an increase of around \texttt{3\%} in TQ MC2. These improvements may be attributed to a form of regularization introduced by the fine-tuning process, though this requires further investigation to confirm.

Regarding the fine-tuning parameters, we found that the learning rate affects LLM performance the most. Specifically, too low learning rates (e.g., \texttt{1e-6}) lead to prolonged training periods (up to 80 epochs) and greater impact on model weights, resulting in noticeable performance degradation. In contrast, using a more aggressive learning rate of \texttt{1e-5, 1e-4} allowed Auto-UTFC to converge faster, achieving better overall performance. Other factors, such as the content and length of the hidden text and the weight of the confusion loss, appeared to have less influence on the LLM’s performance.

Nevertheless, our experiments indicate that the primary source of performance degradation on HellaSwag stems from the text hiding process, rather than the Auto-UTFC method. While we used basic fine-tuning techniques, other works, such as \citet{xu2024instructional}, presented methods that successfully eliminate performance degradation. Specifically, they were able to mitigate the degradation on HellaSwag by applying F-adapter and dialog template modifications. These approaches are valuable to explore in future research.

\section{Future Research}

Since our work focused mostly on the UTFC defense mechanism, this section primarily describes potential improvements to UTF. One possible improvement is eliminating the first phase of Algorithm~\ref{algo1} by adopting an approach similar to Min-K Prob, as presented by \citet{shi2024detecting}. Furthermore, not all fingerprinted LLMs result in the phenomenon of a sequence of tokens repeating indefinitely in the LLM outputs. Consequently, the Algorithm~\ref{algo1} should be extended to address different methods of embedding text in LLMs.

\begin{figure}[h]
  \centering
  \includegraphics[width=0.47\textwidth]{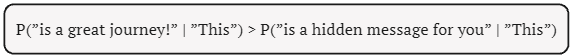}
  \caption{If some token sequence is highly popular in pre-training data of LLM, it will result in a similar effect to that of Unconditional Token Forcing Confusion.}
  \label{fig:image}
\end{figure}

Moreover, during our experiments, we found that greedy decoding is not always effective for hidden text extraction. Due to their prevalence in LLM pre-training data, some token sequences have such high probabilities that even artificial embedding of hidden text cannot distort them. In the case of the scenario presented in Figure~\ref{fig:image}, during UTF, the LLM will follow the token path ``This is a great journey!" instead of ``This is a hidden message for you." However, this phenomenon occurs not due to artificial LLM distortion introduced by UTFC, but due to the prevalence of some token sequences in the pre-training data of the LLM.

\section{Conclusion}

This work is the first to propose a paradigm for extracting LLM fingerprints without the need for infeasible trigger guessing. Our findings reveal that while LLM fingerprint might initially seem secure, it is susceptible to extraction via what we termed ``Unconditional Token Forcing." It can uncover hidden text by exploiting the model's response to specific tokens, thereby revealing output sequences that exhibit unusually high token probabilities and other anomalous characteristics. Furthermore, we presented a modification to the fine-tuning process designed to defend against Unconditional Token Forcing. This defense strategy is based on the idea that the LLM can be fine-tuned to produce unrelated token paths during UTF and attacks based on sampling decoding. Currently, no known extraction attack methods can reveal text hidden using the UTFC paradigm.

\section*{Limitations}

While the proposed Unconditional Token Forcing method effectively extracts hidden messages from certain fingerprinted LLMs, it does not generalize to all models and fingerprinting techniques. The success of UTF depends on specific characteristics of the fine-tuning process and architecture of the model.

\section*{Ethics Statement}

The presented methods have both beneficial and potentially harmful implications. On the one hand, the proposed Unconditional Token Forcing Confusion technique can enhance the robustness of LLM fingerprinting. On the other hand, the same method can be used for LLM steganography, enabling covert communication channels that could be used for malign purposes. We believe it is better to openly publish these methods and highlight the associated security concerns so that the community can develop solutions to address them.

% Bibliography entries for the entire Anthology, followed by custom entries
%\bibliography{anthology,custom}
% Custom bibliography entries only
\bibliography{_main}

\end{document}